\documentclass[
]{ceurart}

\usepackage{listings}
\usepackage{xcolor}
\begin{document}

\copyrightyear{2021}
\copyrightclause{Copyright for this paper by its authors.
  Use permitted under Creative Commons License Attribution 4.0
  International (CC BY 4.0).}

\conference{This manuscript is a preprint and is currently under review.}
\title{PHAX: A Structured Argumentation Framework for User-Centered Explainable AI in Public Health and Biomedical Sciences}

\author[1]{Bahar {\.I}lgen}[%
orcid=0000-0001-5725-0850,
email=ilgenb@rki.de,
]
\address[1]{Center for Artificial Intelligence in Public Health Research (ZKI-PH), Robert Koch-Institut, Nordufer 20, Berlin, 13353, Germany}
\address[2]{Department of Mathematics and Computer Science, Freie Universität Berlin, Arnimallee 14, Berlin, 14195, Germany}

\author[1,2]{Akshat Dubey}[%
orcid=0009-0008-4823-9375,
email=dubeya@rki.de,
]
\author[1,2]{Georges Hattab}[%
orcid=0000-0003-4168-8254,
email=hattabg@rki.de,
]

\begin{abstract}
   Ensuring transparency and trust in AI-driven public health and biomedical sciences systems requires more than accurate predictions—it demands explanations that are clear, contextual, and socially accountable. While explainable AI (XAI) has advanced in areas like feature attribution and model interpretability, most methods still lack the structure and adaptability needed for diverse health stakeholders, including clinicians, policymakers, and the general public. We introduce PHAX—a Public Health Argumentation and eXplainability framework—that leverages structured argumentation to generate human-centered explanations for AI outputs. PHAX is a multi-layer architecture combining defeasible reasoning, adaptive natural language techniques, and user modeling to produce context-aware, audience-specific justifications.  More specifically, we show how argumentation enhances explainability by supporting AI-driven decision-making, justifying recommendations, and enabling interactive dialogues across user types. We demonstrate the applicability of PHAX through use cases such as medical term simplification, patient-clinician communication, and policy justification. In particular, we show how simplification decisions can be modeled as argument chains and personalized based on user expertise—enhancing both interpretability and trust. By aligning formal reasoning methods with communicative demands, PHAX contributes to a broader vision of transparent, human-centered AI in public health.
\end{abstract}

\begin{keywords}
  Explainable AI \sep
  Argumentative Explanation \sep
  Structured Argumentation \sep
  User-Adaptive Explanation  \sep
  Public Health Informatics \sep
  Biomedical NLP
\end{keywords}

\maketitle

\section{Introduction}

   As artificial intelligence (AI) becomes increasingly embedded in public health systems, ensuring that AI outputs are understandable, trustworthy, and tailored to diverse stakeholders has become a critical challenge \cite{topol2019, wiens2019, amann2020,  hattab2025ai, ploug2021, dubey2024}. Moreover, recent calls in public health literature highlight the necessity of Explainable AI (XAI) to foster transparency and professional trust in healthcare applications. From clinical diagnostics to vaccination policy, AI now plays a role in high-stakes decisions that affect patients, practitioners, and entire populations. Recent applications in public health, such as those focused on pandemic preparedness, have further demonstrated the need for XAI in epidemiological decision-making contexts \cite{khalili2024}. Despite such promising directions, the reasoning behind many AI-driven decisions remains opaque, raising serious concerns about accountability, equity, and interpretability.

   XAI seeks to address these concerns by making model behavior more transparent. However, most existing XAI approaches—such as feature attribution or counterfactual analysis—struggle to provide user-adaptive and communicatively effective explanations, especially in language-based applications \cite{mindlin2024, he2025}. Such limitations are particularly problematic in public health and biomedical contexts, where information must be both technically accurate and socially communicable. Recent work in human-computer interaction (HCI) has also emphasized the need for explainable, accountable, and intelligible systems~\cite{abdul2018}. Taken together, these challenges call for a new paradigm in explainability that mirrors how humans reason and justify decisions. In this context, explanation should be modeled as a reasoning process, not just a visualization or annotation. 
   
   Defining what constitutes an explanation is itself a complex issue. As reviewed in~\cite{miller2019}, explanations have been conceptualized in various ways: as assignments of causal responsibility~\cite{josephson1996}, as both the process and product of addressing a "Why?" question~\cite{lombrozo2006}, and as a means of constructing shared meaning. These perspectives highlight that explanation is not merely a factual output but a communicative and cognitive process that engages reasoning and interpretation.
   To this end, we propose PHAX: a Public Health Argumentation and eXplainability framework. PHAX is a multi-layer architecture integrating structured argumentation, adaptive natural language processing (NLP), and user modeling to generate clear, audience-specific justifications for AI outputs. It treats explanation not as a post-hoc add-on, but as a first-class component of decision-making pipelines. Structured argumentation, in particular, serves as a foundational mechanism: it enables AI systems to articulate decision processes step by step, manage uncertainty, and resolve conflicting evidence through formally grounded reasoning~\cite{vassiliades2021}. Such capabilities are essential for building trust in AI-driven public health systems and biomedical systems. More specifically, within the domains of public health and biomedical sciences, we demonstrate how argumentation enhances explainability, with applications spanning areas such as decision-making (e.g., vaccination prioritization or clinical risk stratification), justification of system outputs (e.g., medical term simplification or the selection of diagnostic biomarkers), and interactive dialogue (e.g., clinician-AI interaction in diagnosis or treatment planning). These capabilities allow AI systems to deliver context-sensitive explanations aligned with stakeholder needs across both population-level and individual-level biomedical applications. Through structured reasoning and audience-aware communication, argumentation enables AI systems to provide transparent, context-sensitive explanations across a range of high-stakes scenarios in public health and biomedical sciences.

   PHAX builds on the formal tools of argumentation theory—including Dung's Abstract Framework \cite{dung1995} and ASPIC$^+$—to model \cite{modgil2013} outputs as defeasible claims supported by reasoning chains. It also incorporates adaptive NLP techniques such as text simplification (TS), semantic role labeling (SRL), and discourse parsing, and audience-aware surface realization to tailor explanations to different users. Whether the audience is a patient, clinician, or policymaker, PHAX generates logically grounded and socially appropriate explanations.

   To demonstrate the utility of PHAX, we present medical text simplification (MTS) as a core use case. Simplification decisions—such as replacing "myocardial infarction" with "heart attack" — are modeled as arguments, based on corpus frequency, semantic equivalence, and contextual appropriateness. Explanations are then adjusted in tone and depth based on user profiles. This showcases how PHAX enhances interpretability, transparency, and trust in a critical public health application.
   This paper makes the following contributions:
   \begin{itemize}
       \item Introduces PHAX, a novel framework that integrates structured argumentation and adaptive NLP for explainable AI in public health and biomedical sciences.
       \item Demonstrates how simplification and other AI outputs can be modeled as defeasible reasoning chains.
       \item Proposes user-adaptive explanation strategies tailored to different stakeholders.
       \item Provides illustrative use cases to highlight PHAX's applicability in diverse public health contexts.
   \end{itemize}

\section{Related Work}

   XAI encompasses a range of approaches designed to make model behavior more interpretable. Common techniques include feature attribution methods (e.g., LIME, SHAP), saliency mapping, and counterfactual reasoning. These methods aim to provide insight into how AI models arrive at their predictions, but they often lack the ability to produce explanations that are user-adaptive and socially contextualized—particularly in domains like public health and biomedical sciences~\cite{dwivedi2023}.

   Following the limitations of purely statistical or post-hoc methods, recent research has turned to structured argumentation as a foundation for explanation in AI systems. Frameworks based on Dung’s Abstract Argumentation Framework (AF) and ASPIC$^+$ have been explored as mechanisms to model reasoning processes and support step-by-step justifications for AI outputs. For instance, Vassiliades et al.\cite{vassiliades2021} and Čyras et al.\cite{cyras2021} survey a range of argumentation-based XAI approaches, showing how argument structures can provide more transparent and logically grounded explanations, particularly in settings involving uncertainty or conflicting information.

   While structured argumentation has provided a solid foundation for explainable AI, most existing approaches emphasize formal rigor over communicative utility. Prior work has largely focused on symbolic representations and internally consistent reasoning frameworks, often overlooking how explanations are interpreted, contextualized, or acted upon by diverse end-users. To address these limitations, the PHAX framework proposed in this work builds on argumentation theory but goes beyond structural clarity by integrating user modeling and adaptive natural language generation. Unlike prior approaches, PHAX aims to deliver context-sensitive, stakeholder-specific justifications that are not only logically coherent but also socially meaningful.
   
   Several applications of argumentation-based explainability have emerged in the biomedical and healthcare domains. Argumentation theory has been shown to support clinical decision-making under uncertainty and incomplete information. Longo et al.~\cite{longo2012} demonstrate how defeasible reasoning and formal argumentation frameworks can model expert reasoning in cancer recurrence prediction. This aligns with broader research in hybrid intelligence, which emphasizes collaborative, explainable AI systems that support human reasoning rather than replace it~\cite{akata2020}. This perspective underscores the role of argumentation and explanation as core components in designing trustworthy and transparent systems, especially in high-stakes domains such as healthcare. Such approaches stress the importance of aligning machine reasoning with human cognitive and ethical expectations—an objective well supported by structured argumentation frameworks such as PHAX. The integration of user-adaptive justifications and formal inference mechanisms in PHAX contributes to this vision by ensuring that explanations remain both logically sound and socially intelligible across diverse health contexts.
   
   One concrete example of such a system is the CONSULT project~\cite{kokciyan2019}, which applies computational argumentation to clinical settings. CONSULT combines data from EHRs, wearable sensors, and treatment guidelines to support collaborative decision-making. It leverages ASPIC$^+$ for reasoning under uncertainty and generates argumentation-based dialogues to explain treatment recommendations to patients and healthcare professionals. CONSULT's use of argument schemes, attack relations, and user-facing explanations closely aligns with the goals of the PHAX framework in structuring and communicating personalized justifications across stakeholder groups.

\section{PHAX: Public Health Argumentation and eXplainability Framework}
\subsection{Architecture and Layers}
\begin{figure}
  \centering
  \includegraphics[width=\linewidth]{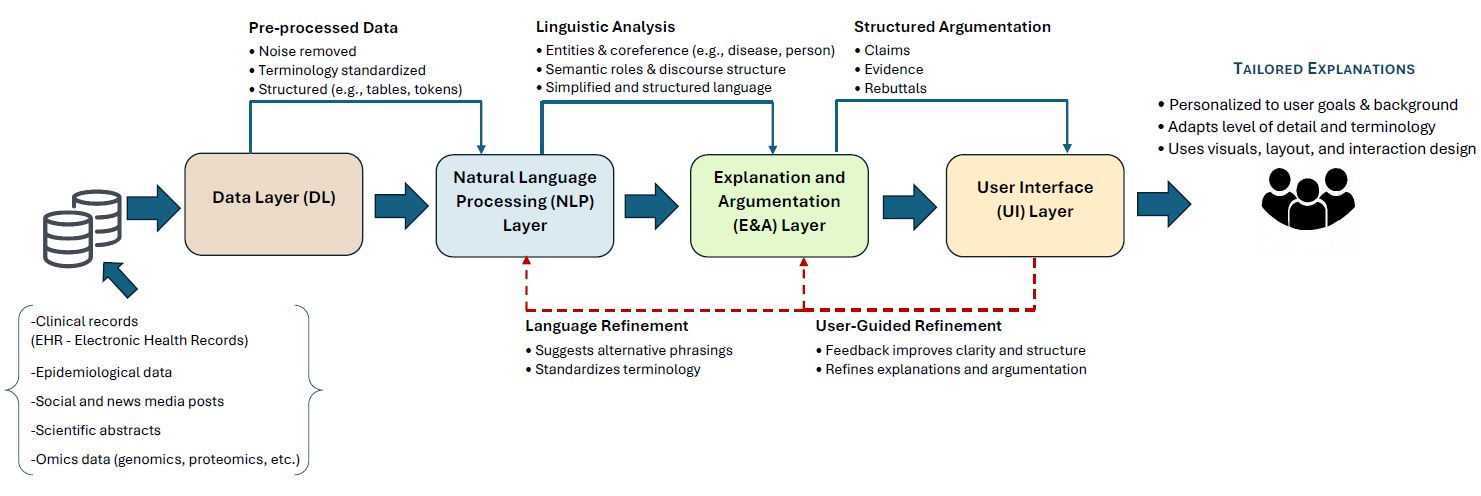}
  \caption{The PHAX layered architecture for user-centered explainable AI in public health.}
  \label{fig:phax_architecture}
\end{figure}

   PHAX (Public Health Argumentation and eXplainability) is a structured argumentation framework designed to enhance the transparency, accountability, and user alignment of AI systems in public health and biomedical domains. By embedding explainability within reasoning structures, it addresses limitations of post-hoc or model-agnostic XAI approaches in complex, high-stakes settings. It integrates formal reasoning models, NLP, and audience-aware explanation strategies to generate context-sensitive, socially meaningful justifications for AI outputs. Unlike conventional explainability methods that treat explanation as a post-hoc step, PHAX embeds explanation generation directly into the AI decision-making pipeline. This allows the system not only to justify its decisions but also to adapt them based on counterarguments, uncertainty, and user-specific needs. Table 1 outlines how PHAX operationalizes key XAI goals \cite{vassiliades2021} through NLP tasks in the context of public health applications. The overall architecture of PHAX is structured into four main layers, progressing from raw data to user-adaptive explanation. As illustrated in Figure~\ref{fig:phax_architecture}, each layer transforms and passes information to the next, with user feedback loops enabling refinement at multiple stages. Argumentation serves as the core reasoning mechanism that translates NLP-derived insights into structured justifications. It connects the output of the NLP Layer to both the internal logic of decision-making and the external communicative needs of the user interface.

    \begin{itemize}
        \item \textbf{Data Layer:} This layer collects and preprocesses heterogeneous data sources -including clinical texts, patient records, epidemiological databases, and social media content- for downstream processing. It ensures that structured and unstructured sources are harmonized before reasoning begins.
        \item \textbf{NLP Processing Layer:} This component performs domain-specific language analysis through tasks such as named entity recognition (NER), semantic role labeling (SRL), discourse parsing, and text simplification. Its outputs provide structured input for argument construction and semantic validation.
        \item \textbf{Explanation and Argumentation Layer:} At this level, system outputs are modeled as defeasible arguments using formal structures like Dung’s Abstract Argumentation Framework and ASPIC+. Arguments are built around claims (e.g., a proposed simplification), supports (e.g., corpus frequency or semantic equivalence), and counterarguments (e.g., ambiguity or domain-specific concerns). This layer formalizes reasoning steps and enables the system to manage uncertainty and conflicting evidence.
        \item \textbf{User Interface Layer:} Finally, this layer delivers the generated explanations through interfaces such as dashboards, conversational agents, or textual summaries. It adapts the explanation’s tone, structure, and depth depending on the user type—whether patient, clinician, or policymaker. It closes the loop between formal reasoning and human interpretability.
    \end{itemize}

     While traditional explainability frameworks separate development-time and deployment-time tasks, PHAX is designed with a layered architecture in which explanation-relevant processes are continuously integrated throughout the AI system’s lifecycle. Each layer contributes both to model construction and real-time explanation generation, facilitating continuous traceability and stakeholder alignment.

\begin{table}[h]
\centering
\caption{Mapping XAI Objectives to NLP Tasks in the PHAX Framework}
\begin{tabular}{
  >{\raggedright\arraybackslash}p{2.7cm}|
  >{\raggedright\arraybackslash}p{4.4cm}|
  >{\raggedright\arraybackslash}p{6cm}}
\hline
\textbf{XAI Objective} & \textbf{NLP Task} & \textbf{Illustrative Use Case} \\
\hline
Transparency & Discourse Parsing, \makecell{Semantic Role Labeling} & Explaining vaccine recommendation steps in a logical sequence \\
Justification & Argumentation Mining, \makecell{Natural Language Inference} & Providing structured evidence for prioritizing vaccination groups \\
Relevance & Question Answering, \makecell{Information~Retrieval} & Answering “Why are masks still needed?” with data-backed explanations \\
Conceptualization & Text Simplification, \makecell{Named Entity Recognition} & Simplifying terms like “PCR Test” for lay audiences \\
Learning & Dialogue Systems, \makecell{Knowledge Graphs} & Teaching how vaccines work through chatbot interactions \\
\hline
\end{tabular}
\label{tab:xai_nlp}
\end{table}

\subsection{Common Argumentation Schemes in Public Health and Biomedical Reasoning}

   In public health and biomedical domains, decisions often involve uncertain data, multiple stakeholders, ethical constraints, and context-sensitive reasoning. Different types of argumentation structures—such as causal, analogical, practical, or expert-driven—are used depending on the decision type and audience.

   PHAX leverages well-established argumentation schemes to generate explanations that are both structured and intuitive to human reasoning. Argumentation schemes are stereotypical patterns of reasoning that capture how people commonly justify claims in different contexts~\cite{walton2008argumentation}. Each scheme defines a type of inference (e.g., from expert authority or practical goals) along with typical critical questions used to challenge it.

   In the domains of public health and biomedical decision-making, such schemes provide an effective foundation for crafting user-facing justifications. Table~\ref{tab:schemes} lists several commonly used schemes, adapted for relevant real-world scenarios. These are not only useful for structuring logical support but also serve as templates for natural language explanations that align with stakeholder expectations.

\begin{table}[ht]
\centering
\caption{Argumentation Schemes (AS)s in Public Health and Biomedical Reasoning}
\label{tab:schemes}
\begin{tabular}{p{3.6cm}|p{4.6cm}|p{5.5cm}}
\hline
\textbf{AS} & \textbf{Description} & \textbf{Example in Context} \\
\hline
Expert Opinion & Relying on authority or professional expertise & ``WHO recommends vaccination for this age group'' \\

Cause to Effect & Predicting consequences of an action or event & ``Masking reduces viral transmission'' \\

Practical Reasoning & Choosing actions to achieve desired outcomes & ``To prevent ICU overload, implement lockdown'' \\

Analogy & Inferring based on similarity to previous~cases & ``Contact tracing worked for Ebola; it can help for COVID'' \\

Statistical Generalization & Drawing conclusions from population-level data & ``This drug helped 70\% of patients in clinical trials'' \\

Ethical/Value-based & Arguing based on fairness, harm, or social values & ``We must prioritize vulnerable groups to ensure equity'' \\
\hline
\end{tabular}
\end{table}

\subsubsection*{Formal Representation of Schemes}
PHAX uses formal representations of argumentation schemes to support logic-based justification and reasoning. Below are selected examples:

\paragraph{Formalization:} The scheme for $Expert Opinion$ where $P$ is the proposition under consideration and $D$ is the relevant domain, can be encoded as:

\[
\text{is\_expert}(E, D),\ \text{asserts}(E, P),\ \text{relevant}(P, D) \Rightarrow \text{believe}(P)
\]

\paragraph{Cause to Effect:}
\[
\text{action}(A),\ \text{causes}(A, E) \Rightarrow \text{expect}(E)
\]

\paragraph{Practical Reasoning:}
\[
\text{goal}(G),\ \text{action}(A),\ \text{promotes}(A, G) \Rightarrow \text{do}(A)
\]

\subsection{Structured Reasoning and Argumentative Explanation in PHAX}

    To support structured and adaptable explanations, PHAX relies on a hybrid formal foundation that combines elements from deductive, structured, and label-based argumentation models. A PHAX builds on structured argumentation principles, enabling precise tracking of how conclusions are derived—essential in biomedical and public health scenarios where traceable reasoning is critical. This is particularly useful in biomedical and public health scenarios, where explanations must be based on traceable and valid reasoning steps.

    The framework builds on formalisms such as ASPIC+, which enables both strict (deductive) and defeasible rules. While strict rules model clear-cut logic (e.g., eligibility based on clinical criteria), defeasible rules allow the system to reason under uncertainty and exceptions—an essential feature in high-stakes public health decision-making.

    Additionally, PHAX adopts principles from label-based argumentation to handle preference, uncertainty, and credibility. Arguments can be labeled with confidence levels, stakeholder relevance, or ethical weights, which are propagated through the argument graph to guide resolution. This makes the system sensitive to contextual and user-specific needs, supporting more personalized and socially attuned justifications.

    Finally, PHAX incorporates argumentation schemes, such as Expert Opinion, Practical Reasoning, and Cause to Effect, which reflect common patterns of human reasoning. These schemes serve as templates to generate natural language explanations that are cognitively aligned with how different stakeholders interpret justification—enhancing both transparency and persuasive power.

    At the core of PHAX is the use of structured argumentation to represent and explain AI-generated outputs. Each decision is modeled as a claim supported by explicit premises and, when appropriate, challenged by potential objections—mirroring human reasoning and enabling transparent justifications. For instance, in medical text simplification, PHAX treats the decision to simplify a term $X$ to $Y$ not merely as output, but as an argument that can be analyzed and, if needed, contested.

\subsubsection{Illustrative Example: Medical Simplification as Structured Argument}
\begin{itemize}
    \item \textbf{Claim:} Term $X$ is simplified to $Y$.
    \item \textbf{Support 1:} $Y$ is more frequent in layperson corpora.
    \item \textbf{Support 2:} No semantic loss is detected via a Natural Language Inference (NLI) model.
    \item \textbf{Attack:} $Y$ may be ambiguous in clinical contexts.
\end{itemize}

   These structures are formalized using ASPIC$^+$, enabling both graphical visualization and logic-based evaluation. This approach goes beyond surface-level explainability by exposing the reasoning process itself. In particular, the decision to simplify a term $X$ to $Y$ is not presented as a final output alone, but accompanied by explicit justifications and possible objections. This aligns with the principles of structured argumentation used in explainable AI.

\subsubsection*{Dung’s Abstract Argumentation Framework (AF)}

In Dung’s AF~\cite{dung1995}, arguments are modeled as atomic elements with defined attack relations.

Let:
\begin{align*}
    A &: \text{The argument supporting the simplification of $X \rightarrow Y$} \\
    B &: \text{Support based on frequency: "$Y$ is more frequent than $X$"} \\
    C &: \text{Support based on semantic similarity: "No meaning lost"} \\
    D &: \text{Counterargument: "$Y$ is ambiguous in clinical contexts"}
\end{align*}

We define the argument set and the attack relation as follows:

\[
\text{Args} = \{A, B, C, D\}, \quad \text{Attacks} = \{(D, A)\}
\]

Here, $D$ challenges the simplification decision, which can be evaluated using grounded or preferred semantics depending on the context.

\subsubsection*{ASPIC$^+$ Representation}

ASPIC$^+$~\cite{modgil2013} enriches this view by including internal structure, rules, and types of reasoning. The same example can be modeled as:

\paragraph{Premises:}
\begin{align*}
    P_1 &: \text{frequency}(Y) > \text{frequency}(X) \\
    P_2 &: \text{semantic\_match}(X, Y) = \text{True} \\
    P_3 &: \text{ambiguity}(Y) = \text{High\_Clinical}
\end{align*}

\paragraph{Defeasible Rules:}
\begin{align*}
    r_1 &: (P_1, P_2) \Rightarrow \text{prefer}(Y) \\
    r_2 &: P_3 \Rightarrow \neg \text{prefer}(Y)
\end{align*}

\paragraph{Arguments:}
\begin{itemize}
    \item $Arg_1 = \langle P_1, P_2, r_1 \rangle \Rightarrow \text{prefer}(Y)$
    \item $Arg_2 = \langle P_3, r_2 \rangle \Rightarrow \neg \text{prefer}(Y)$
\end{itemize}

Here, $Arg_2$ attacks $Arg_1$, resulting in a defeasible justification structure. The system can select or reject the simplification based on external preferences, such as the user's role (e.g., patient or clinician).

This structured approach allows PHAX to generate explainable outputs that go beyond readability scores, instead providing reasoned justifications that can be tailored and interrogated across use cases.

\subsubsection{Formalization: Evidence-Based Reasoning via PICO}

Beyond surface-level linguistic applications such as term simplification, PHAX’s formal reasoning capabilities extend to evidence-based clinical logic. The following illustrates how structured argumentation can be applied to biomedical literature analysis, using the widely adopted PICO paradigm. Building on the earlier simplification use case, we now illustrate how the same argumentation machinery can support evidence-based clinical reasoning through the PICO paradigm. In particular, structured argumentation offers a compelling foundation for modeling \textit{evidence-based claims} derived from biomedical literature using the PICO (Population, Intervention, Comparison, Outcome) paradigm.

Each PICO element can be represented as a formal predicate:
\begin{itemize}
    \item $P(x)$: entity $x$ belongs to the target population  
    \item $I(x)$: intervention applied  
    \item $C(x)$: control or comparison condition  
    \item $O(x)$: observed or expected outcome  
\end{itemize}

These allow the construction of defeasible rules such as:
\[
P(x) \wedge I(x) \Rightarrow_O O(x)
\]
This implies that for individuals in population $P$, the application of intervention $I$ leads to outcome $O$—under typical conditions. However, due to co-morbidities, alternative studies, or contextual constraints, such a rule remains \textit{defeasible}. Counterarguments may cite exceptions (e.g., ``$I$ is contraindicated for subgroups in $P$'').

Using ASPIC$^+$, such clinical evidence can be formalized as follows:

\textbf{Premises:}
\begin{align*}
    P_1 &: \text{The study population matches } P \\
    P_2 &: \text{Intervention } I \text{ was applied} \\
    P_3 &: \text{Outcome } O \text{ was observed} \\
    P_4 &: \text{Source study has high credibility}
\end{align*}

\textbf{Defeasible Rule:}
\[
(P_1, P_2, P_3, P_4) \Rightarrow \text{recommend}(I, P)
\]

\textbf{Counterargument Example:} A second study finds conflicting evidence, yielding:
\[
(P'_1, P'_2, \neg O, P'_4) \Rightarrow \neg \text{recommend}(I, P)
\]

These opposing arguments can then be compared via preference criteria (e.g., study quality, sample size) and evaluated within an argumentation framework using grounded or preferred semantics. Grounded semantics selects the most cautious acceptable set of arguments, while preferred semantics favors maximal admissible sets.

This formalism not only enhances interpretability of AI recommendations in public health contexts, but also allows \textbf{systematic traceability} of how and why a certain intervention is proposed—bridging evidence-based medicine and explainable AI.

\subsection{User-Adaptive Explanation Generation}

Public health communication involves a range of stakeholders—clinicians, policymakers, patients—each with different cognitive needs and expectations. To support effective communication, PHAX dynamically adapts both the structure and the presentation of its explanations based on user profiles. These user modeling attributes—such as expertise, lexical tolerance, and cognitive expectations—govern how explanations are tailored across multiple adaptation layers, as illustrated below.
\paragraph{Theoretical Foundation.}
Drawing from Relevance Theory~\cite{sperber1986} and Grice's Cooperative Principles~\cite{grice1975}, PHAX ensures that explanations are not only accurate but also cognitively appropriate for the intended audience. This is operationalized through user modeling and selective generation of explanation content. Furthermore, PHAX incorporates principles from Labelled Argumentation Frameworks \cite{amgoud2013ranking,hunter2013} to propagate metadata such as confidence, role-based preference, or ethical weight across the explanation graph.

\paragraph{Definition 1. (User Profile)}
A user profile \( U \) is a tuple \( (e, l, c) \), where:
\begin{itemize}
  \item \( e \in \mathbb{R} \): Domain expertise level (e.g., clinician vs. layperson)
  \item \( l \in \mathbb{R} \): Lexical tolerance (e.g., jargon sensitivity)
  \item \( c \in \mathbb{R} \): Cognitive depth (e.g., expected explanation complexity)
\end{itemize}

\paragraph{Definition 2. (Semantic Sufficiency)}
Given an explanation tree \( T \) and argument \( a \), semantic sufficiency \( \sigma_T(a) \in [0,1] \) quantifies the extent to which \( T \) supports \( a \), possibly via aggregation over leaf node support and edge weights.

\paragraph{Definition 3. (Utility Function)}
Utility is a linear combination of weighted factors:
\[
\text{Utility}(T, U) = \sum_{i=1}^n w_i \cdot f_i(T, U)
\]
where \( f_i \) is a feature function (e.g., clarity, lexical fit), and \( w_i \in \mathbb{R} \) is a tunable weight.

\paragraph{Formal Mechanism.}
Each user is modeled as a profile \( U \) with attributes including domain expertise, lexical tolerance, and cognitive depth. Given a full Quantitative Dispute Tree \( QDT(a) \) \cite{cyras2022qdt} for an argument \( a \), the framework selects a user-appropriate subgraph \( T^* \) as follows:

\[
T^* = \arg\max_T \text{Utility}(T, U) \quad \text{subject to} \quad \sigma_T(a) \geq \tau(a)
\]

\[
\text{Utility}(T, U) = \alpha \cdot \text{Clarity}(T, U) + \beta \cdot \text{Relevance}(T, U) + \gamma \cdot \text{LexicalFit}(T, U)
\]

Where:
\begin{itemize}
    \item \( \sigma_T(a) \): Semantic sufficiency — does \( T \) still justify argument \( a \)?
    \item \( \tau(a) \): Task-defined threshold for completeness
\end{itemize}

\paragraph{Adaptation Dimensions.}
\begin{itemize}
    \item \textbf{Lexical complexity:} Simplified phrasing and terminology for lay users
    \item \textbf{Information depth:} Full argument chains for experts, summarized reasoning for general audiences
    \item \textbf{Presentation format:} Visualizations for policymakers, natural language for patients, dialogue-based interaction for clinicians
\end{itemize}

\paragraph{Illustrative Example.}
A recommendation to prioritize vaccine distribution may be explained differently:

\begin{itemize}
    \item \textbf{To a clinician:} “This decision is supported by Phase III trial data showing 92\% efficacy.”
    \item \textbf{To a patient:} “This vaccine has helped many people like you stay safe.”
    \item \textbf{To a policymaker:} “Prioritizing this group prevents overloading ICUs by 45\%.”
\end{itemize}

\paragraph{Relation to Argumentation Schemes.}
User-tailored explanations are grounded in different argumentation schemes depending on audience expectations:
\begin{itemize}
    \item \textit{Cause to Effect} — For lay users: “Vaccination reduces risk of severe disease.”
    \item \textit{Statistical Generalization} — For domain experts: “70\% of patients showed improvement.”
    \item \textit{Practical Reasoning} — For decision-makers: “To prevent ICU overload, prioritize group A.”
    \item \textit{Ethical Reasoning} — For public discourse: “We must protect the most vulnerable first.”
\end{itemize}

\paragraph{Connection to User Interface Layer.}
These adaptive explanations are operationalized through the User Interface Layer of PHAX, which selects and renders the appropriate format and depth of explanation based on the computed utility for each user profile. Whether delivered as textual justifications, interactive dialogues, or visual dashboards, the UI layer serves as the final conduit through which tailored argument structures are communicated. This ensures that the formal reasoning modeled in earlier layers is not only preserved but made accessible and persuasive for its intended audience.

\section{Application Scenarios Across Public Health and Biomedical Sciences}

PHAX addresses a broad spectrum of reasoning and communication challenges in public health and biomedical domains, where decisions often involve uncertainty, competing values, and diverse stakeholders. Beyond its core architecture, the framework provides structured and audience-sensitive explanations tailored to real-world needs—from clinical decision support to public communication. Below, we present representative scenarios illustrating how PHAX integrates argumentation and explanation to promote transparency, trust, and actionable insight across practical settings.

\subsection{Decision Support and Stakeholder Alignment}

Public health decisions often involve navigating trade-offs among competing priorities, limited resources, and uncertain outcomes. For example, vaccination prioritization during a pandemic requires balancing exposure risk, equity, and healthcare system capacity. PHAX models such dilemmas using defeasible argumentation, enabling transparent, traceable justifications for complex decisions. Its layered architecture delivers explanations tailored to different stakeholders: clinicians may explore structured evidence trails via interactive dashboards, while policymakers access high-level summaries that emphasize societal trade-offs and ethical considerations.

\subsection{Evidence Synthesis and Biomedical Summarization}

Systematic reviews are essential in biomedical research for aggregating findings across multiple studies, yet their length and heterogeneity often limit accessibility and interpretability. PHAX facilitates structured summarization by applying argumentation mining techniques to PICO-extracted data, capturing key claims, counterclaims, and the credibility of supporting evidence. These elements are formalized into argument structures, enabling contrastive summaries that clarify points of agreement, contention, and uncertainty across studies. Such summaries support clinicians and researchers in efficiently navigating complex and sometimes conflicting bodies of literature.

\subsection{Public Communication and Policy Justification}

Effective public communication of health interventions—such as lockdowns or vaccine mandates—requires a delicate balance between scientific accuracy and audience accessibility. PHAX addresses this challenge by generating explanations grounded in established argumentation schemes (e.g., causal, ethical, practical), and tailoring their phrasing and framing to specific user profiles. For instance, a lockdown policy may be framed in terms of “transmission control” when addressing clinicians, but emphasized as “protecting the vulnerable” in public-facing messages. This audience-sensitive adaptation enhances clarity and trust without compromising factual integrity.

\subsection{Risk Communication and Misinformation Rebuttals}

Health misinformation often proliferates through arguments that are logically flawed yet emotionally persuasive. PHAX addresses this challenge by generating structured rebuttals: it decomposes misinformation claims into their constituent premises, evaluates their validity, and constructs counterarguments grounded in scientific evidence and tailored to the target audience. For instance, the false claim that “vaccines cause infertility” can be refuted through mechanistic evidence and trial data for clinicians, while lay audiences may receive simpler, empathetically framed responses that emphasize safety and social consensus. This audience-aware rebuttal strategy enhances persuasive effectiveness without compromising scientific rigor.

\subsection{Interface-Driven Personalization and Delivery}

The impact of an explanation often depends not only on its content, but also on how it is delivered. PHAX supports multi-modal explanation delivery through interfaces that adapt to users' preferences, literacy levels, and interaction contexts. These include natural language narratives, visual summaries, and dialogue-based interactions. For example, patients may receive conversational explanations via chatbot interfaces, while policymakers might explore comparative scenario graphs that highlight trade-offs. These modalities are selected dynamically based on user modeling, ensuring that the explanation aligns with the user's cognitive and informational needs as captured by PHAX’s adaptive layer.

\section{Conclusion and Future Work}

This study presents PHAX—a Public Health Argumentation and eXplainability framework—designed to support transparent, context-aware, and user-adaptive explanations in high-stakes domains such as healthcare and biomedical sciences. Building upon structured argumentation theory, PHAX incorporates formal reasoning, adaptive NLP pipelines, and user modeling to generate stakeholder-specific justifications for AI outputs. Our contributions include a modular architecture for integrating explainability into the AI lifecycle, a formalization of user-adaptive explanation generation, and illustrative applications in medical term simplification, policy justification, and systematic review summarization. By combining defeasible reasoning, argumentation schemes, and multimodal delivery interfaces, PHAX enables interpretable decision support tailored to diverse user needs.

Looking ahead, we aim to extend PHAX’s reasoning capabilities by incorporating more explicit support for uncertainty-aware and value-sensitive argumentation. These enhancements will better capture the complexity of real-world public health decisions, where conflicting priorities and incomplete information are common. A key direction involves activating PHAX’s adaptive layer through live user feedback, enabling continuous refinement of explanations aligned with user profiles. We also plan to evaluate PHAX in real-world settings through user studies with clinicians, policymakers, and patients, to assess explanation effectiveness, trust calibration, and usability in practice.

\begin{acknowledgments}
This work has been financially supported by the German Federal Ministry of Health (BMG) under grant No.: ZMI5- 2523GHP027 (project “Strengthening National Immunization Technical Advisory Groups and their Evidence-based Decision-making in the WHO European Region and Globally” SENSE) part of the Global Health Protection Programme, GHPP.
\end{acknowledgments}


\bibliography{sample-ceur}

\begin{thebibliography}{27}
\expandafter\ifx\csname natexlab\endcsname\relax\def\natexlab#1{#1}\fi
\providecommand{\url}[1]{\texttt{#1}}
\providecommand{\href}[2]{#2}
\providecommand{\path}[1]{#1}
\providecommand{\DOIprefix}{doi:}
\providecommand{\ArXivprefix}{arXiv:}
\providecommand{\URLprefix}{URL: }
\providecommand{\Pubmedprefix}{pmid:}
\providecommand{\doi}[1]{\href{http://dx.doi.org/#1}{\path{#1}}}
\providecommand{\Pubmed}[1]{\href{pmid:#1}{\path{#1}}}
\providecommand{\bibinfo}[2]{#2}
\ifx\xfnm\relax \def\xfnm[#1]{\unskip,\space#1}\fi
\bibitem[{Topol(2019)}]{topol2019}
\bibinfo{author}{E.~J. Topol},
\newblock \bibinfo{title}{High-performance medicine: The convergence of human and artificial intelligence},
\newblock \bibinfo{journal}{Nature Medicine} \bibinfo{volume}{25} (\bibinfo{year}{2019}) \bibinfo{pages}{44--56}. \DOIprefix\doi{10.1038/s41591-018-0300-7}.
\bibitem[{Wiens et~al.(2019)Wiens, Saria, Sendak, Ghassemi, Liu, Doshi-Velez, Jung, Heller, Kale, Saeed, Ossorio, Thadaney-Israni, and Goldenberg}]{wiens2019}
\bibinfo{author}{J.~Wiens}, \bibinfo{author}{S.~Saria}, \bibinfo{author}{M.~Sendak}, \bibinfo{author}{M.~Ghassemi}, \bibinfo{author}{V.~X. Liu}, \bibinfo{author}{F.~Doshi-Velez}, \bibinfo{author}{K.~Jung}, \bibinfo{author}{K.~Heller}, \bibinfo{author}{D.~Kale}, \bibinfo{author}{M.~Saeed}, \bibinfo{author}{P.~N. Ossorio}, \bibinfo{author}{S.~Thadaney-Israni}, \bibinfo{author}{A.~Goldenberg},
\newblock \bibinfo{title}{Do no harm: A roadmap for responsible machine learning for health care},
\newblock \bibinfo{journal}{Nature Medicine} \bibinfo{volume}{25} (\bibinfo{year}{2019}) \bibinfo{pages}{1337--1340}. \DOIprefix\doi{10.1038/s41591-019-0548-6}.
\bibitem[{Amann et~al.(2020)Amann, Blasimme, Vayena, Frey, Madai, and Consortium}]{amann2020}
\bibinfo{author}{J.~Amann}, \bibinfo{author}{A.~Blasimme}, \bibinfo{author}{E.~Vayena}, \bibinfo{author}{D.~Frey}, \bibinfo{author}{V.~I. Madai}, \bibinfo{author}{P.~Consortium},
\newblock \bibinfo{title}{Explainability for artificial intelligence in healthcare: A multidisciplinary perspective},
\newblock \bibinfo{journal}{BMC Medical Informatics and Decision Making} \bibinfo{volume}{20} (\bibinfo{year}{2020}) \bibinfo{pages}{310}. \DOIprefix\doi{10.1186/s12911-020-01332-6}.
\bibitem[{Hattab et~al.(2025)Hattab, Irrgang, Körber, Kühnert, and Ladewig}]{hattab2025ai}
\bibinfo{author}{G.~Hattab}, \bibinfo{author}{C.~Irrgang}, \bibinfo{author}{N.~Körber}, \bibinfo{author}{D.~Kühnert}, \bibinfo{author}{K.~Ladewig},
\newblock \bibinfo{title}{The way forward to embrace artificial intelligence in public health},
\newblock \bibinfo{journal}{American Journal of Public Health} \bibinfo{volume}{115} (\bibinfo{year}{2025}) \bibinfo{pages}{123--128}. \DOIprefix\doi{10.2105/AJPH.2024.307888}.
\bibitem[{Ploug et~al.(2021)Ploug, Sundby, Moeslund, and Holm}]{ploug2021}
\bibinfo{author}{T.~Ploug}, \bibinfo{author}{A.~Sundby}, \bibinfo{author}{T.~B. Moeslund}, \bibinfo{author}{S.~Holm},
\newblock \bibinfo{title}{Population preferences for performance and explainability of artificial intelligence in health care: Choice-based conjoint survey},
\newblock \bibinfo{journal}{Journal of Medical Internet Research} \bibinfo{volume}{23} (\bibinfo{year}{2021}) \bibinfo{pages}{e26611}. \DOIprefix\doi{10.2196/26611}.
\bibitem[{Dubey et~al.(2024)Dubey, Yang, and Hattab}]{dubey2024}
\bibinfo{author}{A.~Dubey}, \bibinfo{author}{Z.~Yang}, \bibinfo{author}{G.~Hattab},
\newblock \bibinfo{title}{A nested model for ai design and validation},
\newblock \bibinfo{journal}{iScience} \bibinfo{volume}{27} (\bibinfo{year}{2024}) \bibinfo{pages}{110603}. \DOIprefix\doi{10.1016/j.isci.2024.110603}.
\bibitem[{Khalili and Wimmer(2024)}]{khalili2024}
\bibinfo{author}{H.~Khalili}, \bibinfo{author}{M.~A. Wimmer},
\newblock \bibinfo{title}{Towards improved xai-based epidemiological research into the next potential pandemic},
\newblock \bibinfo{journal}{Life} \bibinfo{volume}{14} (\bibinfo{year}{2024}) \bibinfo{pages}{783}. \DOIprefix\doi{10.3390/life14070783}.
\bibitem[{Mindlin et~al.(2024)Mindlin, Robrecht, Morasch, and Cimiano}]{mindlin2024}
\bibinfo{author}{D.~Mindlin}, \bibinfo{author}{A.~S. Robrecht}, \bibinfo{author}{M.~Morasch}, \bibinfo{author}{P.~Cimiano},
\newblock \bibinfo{title}{Measuring user understanding in dialogue-based xai systems},
\newblock in: \bibinfo{booktitle}{Proceedings of the 27th European Conference on Artificial Intelligence (ECAI 2024)}, \bibinfo{publisher}{IOS Press}, \bibinfo{year}{2024}, pp. \bibinfo{pages}{1148--1155}.
\bibitem[{He et~al.(2025)He, Aishwarya, and Gadiraju}]{he2025}
\bibinfo{author}{G.~He}, \bibinfo{author}{N.~Aishwarya}, \bibinfo{author}{U.~Gadiraju},
\newblock \bibinfo{title}{Is conversational xai all you need? human-ai decision making with a conversational xai assistant},
\newblock in: \bibinfo{booktitle}{Proceedings of the 30th International Conference on Intelligent User Interfaces (IUI 2025)}, \bibinfo{publisher}{ACM}, \bibinfo{year}{2025}, pp. \bibinfo{pages}{907--924}. \DOIprefix\doi{/10.1145/3708359.3712133}.
\bibitem[{Abdul et~al.(2018)Abdul, Vermeulen, Wang, Lim, and Kankanhalli}]{abdul2018}
\bibinfo{author}{A.~Abdul}, \bibinfo{author}{J.~Vermeulen}, \bibinfo{author}{D.~Wang}, \bibinfo{author}{B.~Y. Lim}, \bibinfo{author}{M.~Kankanhalli},
\newblock \bibinfo{title}{Trends and trajectories for explainable, accountable and intelligible systems: An hci research agenda},
\newblock in: \bibinfo{booktitle}{Proceedings of the 2018 CHI Conference on Human Factors in Computing Systems}, \bibinfo{publisher}{ACM}, \bibinfo{year}{2018}, pp. \bibinfo{pages}{1--18}. \DOIprefix\doi{10.1145/3173574.3174156}.
\bibitem[{Miller(2019)}]{miller2019}
\bibinfo{author}{T.~Miller},
\newblock \bibinfo{title}{Explanation in artificial intelligence: Insights from the social sciences},
\newblock \bibinfo{journal}{Artificial Intelligence} \bibinfo{volume}{267} (\bibinfo{year}{2019}) \bibinfo{pages}{1--38}. \DOIprefix\doi{/10.1016/j.artint.2018.07.007}.
\bibitem[{Josephson and Josephson(1996)}]{josephson1996}
\bibinfo{author}{J.~R. Josephson}, \bibinfo{author}{S.~G. Josephson}, \bibinfo{title}{Abductive Inference: Computation, Philosophy, Technology}, \bibinfo{publisher}{Cambridge University Press}, \bibinfo{year}{1996}.
\bibitem[{Lombrozo(2006)}]{lombrozo2006}
\bibinfo{author}{T.~Lombrozo},
\newblock \bibinfo{title}{The structure and function of explanations},
\newblock \bibinfo{journal}{Trends in Cognitive Sciences} \bibinfo{volume}{10} (\bibinfo{year}{2006}) \bibinfo{pages}{464--470}. \DOIprefix\doi{10.1016/j.tics.2006.08.004}.
\bibitem[{Vassiliades et~al.(2021)Vassiliades, Bassiliades, and Patkos}]{vassiliades2021}
\bibinfo{author}{A.~Vassiliades}, \bibinfo{author}{N.~Bassiliades}, \bibinfo{author}{T.~Patkos},
\newblock \bibinfo{title}{Argumentation and explainable artificial intelligence: a survey},
\newblock \bibinfo{journal}{The Knowledge Engineering Review} \bibinfo{volume}{36} (\bibinfo{year}{2021}) \bibinfo{pages}{e5}. \DOIprefix\doi{/10.1017/S0269888921000011}.
\bibitem[{Dung(1995)}]{dung1995}
\bibinfo{author}{P.~M. Dung},
\newblock \bibinfo{title}{On the acceptability of arguments and its fundamental role in nonmonotonic reasoning, logic programming and n-person games},
\newblock \bibinfo{journal}{Artificial Intelligence} \bibinfo{volume}{77} (\bibinfo{year}{1995}) \bibinfo{pages}{321--357}. \DOIprefix\doi{10.1016/0004-3702(94)00041-X}.
\bibitem[{Modgil and Prakken(2013)}]{modgil2013}
\bibinfo{author}{S.~Modgil}, \bibinfo{author}{H.~Prakken},
\newblock \bibinfo{title}{A general account of argumentation with preferences},
\newblock \bibinfo{journal}{Artificial Intelligence} \bibinfo{volume}{195} (\bibinfo{year}{2013}) \bibinfo{pages}{361--397}. \DOIprefix\doi{10.1016/j.artint.2012.10.008}.
\bibitem[{Dwivedi et~al.(2023)Dwivedi, Dave, Naik, Singhal, Omer, Patel, Qian, Wen, Shah, Morgan, and Ranjan}]{dwivedi2023}
\bibinfo{author}{R.~Dwivedi}, \bibinfo{author}{D.~Dave}, \bibinfo{author}{H.~Naik}, \bibinfo{author}{S.~Singhal}, \bibinfo{author}{R.~Omer}, \bibinfo{author}{P.~Patel}, \bibinfo{author}{B.~Qian}, \bibinfo{author}{Z.~Wen}, \bibinfo{author}{T.~Shah}, \bibinfo{author}{G.~Morgan}, \bibinfo{author}{R.~Ranjan},
\newblock \bibinfo{title}{Explainable ai (xai): Core ideas, techniques, and solutions},
\newblock \bibinfo{journal}{ACM Computing Surveys} \bibinfo{volume}{55} (\bibinfo{year}{2023}) \bibinfo{pages}{1--33}. \DOIprefix\doi{doi.org/10.1145/3561048}.
\bibitem[{Čyras et~al.(2021)Čyras, Rago, Albini, Baroni, and Toni}]{cyras2021}
\bibinfo{author}{K.~Čyras}, \bibinfo{author}{A.~Rago}, \bibinfo{author}{E.~Albini}, \bibinfo{author}{P.~Baroni}, \bibinfo{author}{F.~Toni},
\newblock \bibinfo{title}{Argumentative xai: A survey},
\newblock in: \bibinfo{booktitle}{Proceedings of the Thirtieth International Joint Conference on Artificial Intelligence (IJCAI-21)}, \bibinfo{year}{2021}, pp. \bibinfo{pages}{4392--4399}. \DOIprefix\doi{10.24963/ijcai.2021/600}.
\bibitem[{Longo et~al.(2012)Longo, Kane, and Hederman}]{longo2012}
\bibinfo{author}{L.~Longo}, \bibinfo{author}{B.~Kane}, \bibinfo{author}{L.~Hederman},
\newblock \bibinfo{title}{Argumentation theory in health care},
\newblock in: \bibinfo{booktitle}{Proceedings of the 25th IEEE Symposium on Computer-Based Medical Systems}, \bibinfo{year}{2012}. \DOIprefix\doi{10.1109/CBMS.2012.6266323}.
\bibitem[{Akata et~al.(2020)Akata, Balliet, Dignum, Eiben, ..., and Welling}]{akata2020}
\bibinfo{author}{Z.~Akata}, \bibinfo{author}{D.~Balliet}, \bibinfo{author}{F.~Dignum}, \bibinfo{author}{G.~Eiben}, \bibinfo{author}{...}, \bibinfo{author}{M.~Welling},
\newblock \bibinfo{title}{A research agenda for hybrid intelligence: Augmenting human intellect with collaborative, adaptive, responsible, and explainable artificial intelligence},
\newblock \bibinfo{journal}{Computer} \bibinfo{volume}{53} (\bibinfo{year}{2020}) \bibinfo{pages}{18--28}. \DOIprefix\doi{10.1109/MC.2020.2996587}.
\bibitem[{K{\"o}kciyan et~al.(2019)K{\"o}kciyan, Chapman, Balatsoukas, Sassoon, Essers, Ashworth, Curcin, Modgil, Parsons, and Sklar}]{kokciyan2019}
\bibinfo{author}{N.~K{\"o}kciyan}, \bibinfo{author}{M.~Chapman}, \bibinfo{author}{P.~Balatsoukas}, \bibinfo{author}{I.~Sassoon}, \bibinfo{author}{K.~Essers}, \bibinfo{author}{M.~Ashworth}, \bibinfo{author}{V.~Curcin}, \bibinfo{author}{S.~Modgil}, \bibinfo{author}{S.~Parsons}, \bibinfo{author}{E.~Sklar},
\newblock \bibinfo{title}{A collaborative decision support tool for managing chronic conditions},
\newblock in: \bibinfo{booktitle}{MEDINFO 2019: Health and Wellbeing e-Networks for All}, \bibinfo{organization}{IOS Press}, \bibinfo{year}{2019}, pp. \bibinfo{pages}{644--648}. \DOIprefix\doi{10.3233/SHTI190302}.
\bibitem[{Walton et~al.(2008)Walton, Reed, and Macagno}]{walton2008argumentation}
\bibinfo{author}{D.~Walton}, \bibinfo{author}{C.~Reed}, \bibinfo{author}{F.~Macagno}, \bibinfo{title}{Argumentation Schemes}, \bibinfo{publisher}{Cambridge University Press}, \bibinfo{address}{Cambridge}, \bibinfo{year}{2008}.
\bibitem[{Sperber and Wilson(1986)}]{sperber1986}
\bibinfo{author}{D.~Sperber}, \bibinfo{author}{D.~Wilson}, \bibinfo{title}{Relevance: Communication and Cognition}, volume \bibinfo{volume}{142}, \bibinfo{publisher}{Harvard University Press}, \bibinfo{address}{Cambridge, MA}, \bibinfo{year}{1986}.
\bibitem[{Grice(1975)}]{grice1975}
\bibinfo{author}{H.~P. Grice},
\newblock \bibinfo{title}{Logic and conversation},
\newblock in: \bibinfo{editor}{P.~Cole}, \bibinfo{editor}{J.~L. Morgan} (Eds.), \bibinfo{booktitle}{Syntax and Semantics: Volume 3 - Speech Acts}, \bibinfo{publisher}{Academic Press}, \bibinfo{year}{1975}, pp. \bibinfo{pages}{41--58}.
\bibitem[{Amgoud and Ben-Naim(2013)}]{amgoud2013ranking}
\bibinfo{author}{L.~Amgoud}, \bibinfo{author}{J.~Ben-Naim},
\newblock \bibinfo{title}{Ranking-based semantics for argumentation frameworks},
\newblock in: \bibinfo{booktitle}{Proceedings of the International Conference on Scalable Uncertainty Management (SUM)}, \bibinfo{publisher}{Springer Berlin Heidelberg}, \bibinfo{address}{Berlin, Heidelberg}, \bibinfo{year}{2013}, pp. \bibinfo{pages}{134--147}. \DOIprefix\doi{10.1007/978-3-642-40381-1_11}.
\bibitem[{Hunter(2013)}]{hunter2013}
\bibinfo{author}{A.~Hunter},
\newblock \bibinfo{title}{A probabilistic approach to modelling uncertain logical arguments},
\newblock \bibinfo{journal}{International Journal of Approximate Reasoning} \bibinfo{volume}{54} (\bibinfo{year}{2013}) \bibinfo{pages}{47--81}. \DOIprefix\doi{/10.1016/j.ijar.2012.08.003}.
\bibitem[{Čyras et~al.(2022)Čyras, Kampik, and Weng}]{cyras2022qdt}
\bibinfo{author}{K.~Čyras}, \bibinfo{author}{T.~Kampik}, \bibinfo{author}{Q.~Weng},
\newblock \bibinfo{title}{Dispute trees as explanations in quantitative (bipolar) argumentation},
\newblock in: \bibinfo{booktitle}{Proceedings of the 1st International Workshop on Argumentation for Explainable AI (ArgXAI)}, volume \bibinfo{volume}{3209}, \bibinfo{publisher}{CEUR Workshop Proceedings}, \bibinfo{year}{2022}, pp. \bibinfo{pages}{1--12}. \URLprefix \url{https://ceur-ws.org/Vol-3209/0872.pdf}.

\end{thebibliography}

\appendix

\end{document}